\documentclass[letterpaper, 10 pt, journal, twoside]{IEEEtran}
\usepackage{amsmath} % assumes amsmath package installed
\usepackage{amssymb}  % assumes amsmath package installed
\usepackage{multirow}
\usepackage[pdftex]{graphicx}
\usepackage{graphics} % for pdf, bitmapped graphics files
\usepackage{graphicx}
\usepackage{subfigure}
\usepackage{cases}
\usepackage{array}
\usepackage{cite}
\usepackage{url}
\usepackage{hyperref}
\usepackage{footnote}
\usepackage{threeparttable}
\usepackage{color}
\usepackage{booktabs}
\usepackage{colortbl}
\usepackage{flushend}

\begin{document}
\title{REDE: End-to-end Object 6D Pose Robust Estimation Using Differentiable Outliers Elimination}

\author{Weitong Hua$^{1}$, Zhongxiang Zhou$^{1}$, Jun Wu$^{1}$, Huang Huang$^{2}$, Yue Wang$^{1}$, Rong Xiong$^{1}$%
\thanks{Manuscript received Oct 15th, 2020; Revised Jan 11th, 2021; Accepted Feb 8th, 2021.}%Use only for final RAL version
\thanks{This paper was recommended for publication by Editor Cesar Cadena upon evaluation of the Associate Editor and Reviewers' comments.
This work was supported by the National Nature Science Foundation of China (U1609210) and the Science and Technology Project of Zhejiang Province (Grant No. 2019C01043).}
\thanks{$^{1}$ are with the State Key Laboratory of Industrial Control Technology and Institute of Cyber-Systems and Control, Zhejiang University, Zhejiang, China. $^{2}$ is with Beijing Institute of Control Engineering. Rong Xiong is the corresponding author {\tt\small rxiong@zju.edu.cn} and Yue Wang is the co-corresponding author {\tt\small wangyue@iipc.zju.edu.cn}.}
\thanks{Digital Object Identifier (DOI): 10.1109/LRA.2021.3062304.}%
}

\markboth{IEEE Robotics and Automation Letters. Preprint Version. Accepted Feb, 2021}
{Hua \MakeLowercase{\textit{et al.}}: REDE: End-to-end Object 6D Pose Robust Estimation Using Differentiable Outliers Elimination}

\maketitle

\begin{abstract}
Object 6D pose estimation is a fundamental task in many applications. Conventional methods solve the task by detecting and matching the keypoints, then estimating the pose. Recent efforts bringing deep learning into the problem mainly overcome the vulnerability of conventional methods to environmental variation due to the hand-crafted feature design. However, these methods cannot achieve end-to-end learning and good interpretability at the same time. In this paper, we propose REDE, a novel end-to-end object pose estimator using RGB-D data, which utilizes network for keypoint regression, and a differentiable geometric pose estimator for pose error back-propagation. Besides, to achieve better robustness when outlier keypoint prediction occurs, we further propose a differentiable outliers elimination method that regresses the candidate result and the confidence simultaneously. Via confidence weighted aggregation of multiple candidates, we can reduce the effect from the outliers in the final estimation. Finally, following the conventional method, we apply a learnable refinement process to further improve the estimation. The experimental results on three benchmark datasets show that REDE slightly outperforms the state-of-the-art approaches and is more robust to object occlusion. Our code is available at \href{https://github.com/HuaWeitong/REDE}{\emph{https://github.com/HuaWeitong/REDE}}.
\end{abstract}

% \begin{IEEEkeywords}
% IEEE, IEEEtran, journal, \LaTeX, paper, template.
% \end{IEEEkeywords}

% \fi
\IEEEpeerreviewmaketitle

\begin{figure}[t]
	\centering
	\subfigure[Conventional line.]{
		\includegraphics[width=1\linewidth]{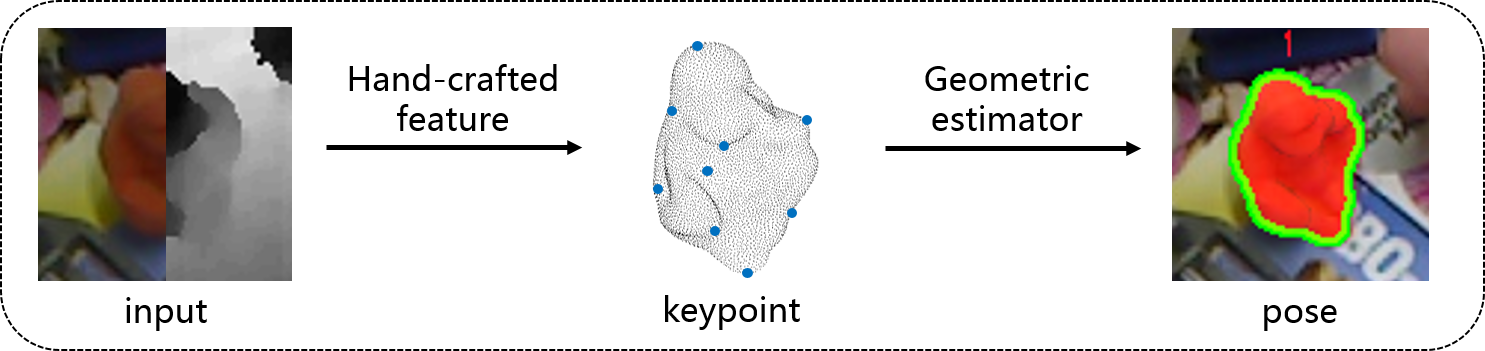}
		%\caption{fig1}
	}
	\subfigure[Direct regression line.]{
		\includegraphics[width=1\linewidth]{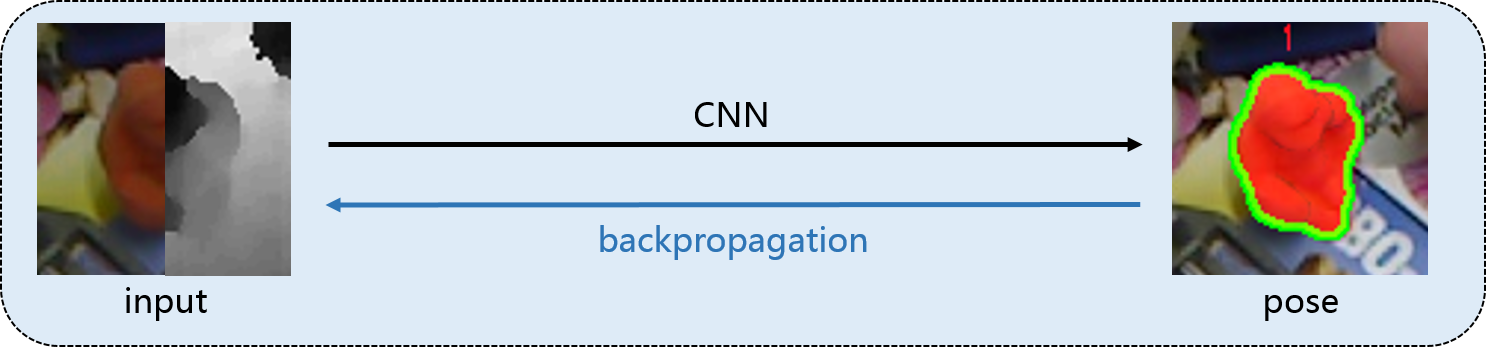}
	}
	\subfigure[Keypoint regression line.]{
		\includegraphics[width=1\linewidth]{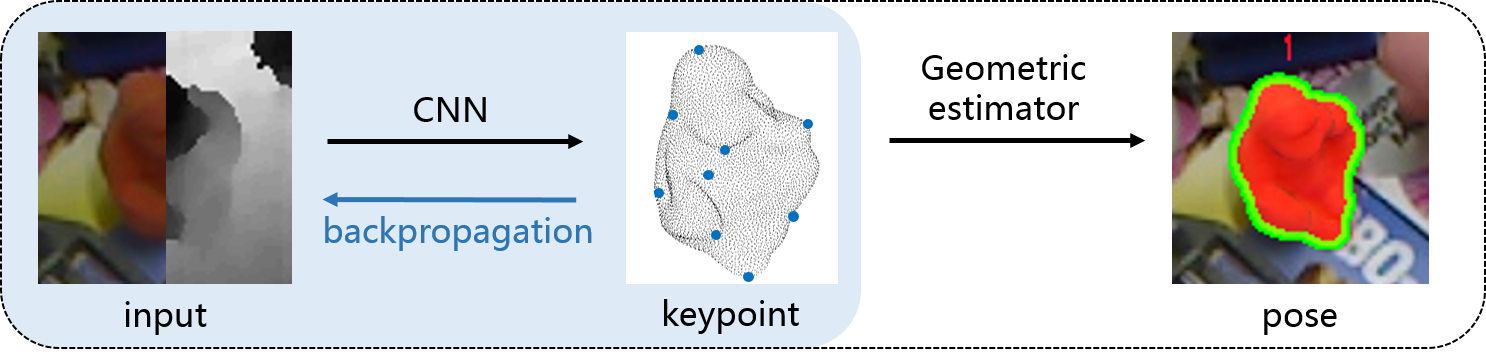}
	}
	\subfigure[REDE.]{
		\includegraphics[width=1\linewidth]{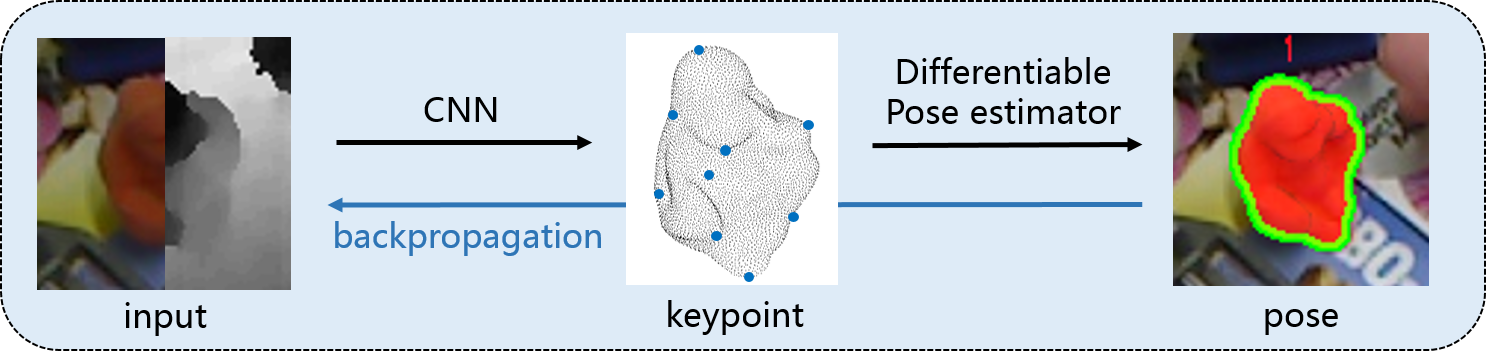}
	}
	\caption{Illustration comparing conventional methods (a), direct regression methods (b), keypoint regression methods (c) and REDE (d). All trainable components are highlighted in blue. REDE integrates the end-to-end learning in \emph{direct regression}, into the \emph{keypoint regression}.}
	\label{fig:method}
	\vspace{-0.4cm}
\end{figure}

\section{Introduction}

\IEEEPARstart{T}{he} task of object 6D pose estimation is to predict the 3D rotation and 3D translation of the object in the current scene with respect to the world coordinates fixed on the object. It is very important in many applications, such as augmented reality \cite{marchand2015pose}, automatic driving \cite{chen2017multi, xu2018pointfusion} and robot grasping \cite{zhu2014single,tremblay2018deep}. Conventionally, a popular method for object pose estimation is built upon hand-crafted keypoint detection and feature matching. The pose is then estimated utilizing the point correspondences between the current image and the object model \cite{drost2010model, hinterstoisser2011multimodal, hinterstoisser2012model}. The main weakness of this pipeline is that the feature design is not learnable to fit the data, leading to difficulties on pose estimation for featureless objects. More recently, with the progress of deep learning, network based pose estimation becomes more popular. There are mainly two lines of works. In the first line, named as \emph{direct regression}, the pose estimation is learned by a fully connected network as a regression task from feature to pose \cite{xiang2017posecnn, do2018deep, wang2019densefusion}, thus achieving the end-to-end learning. However, due to the highly nonlinearity structure in rotation, such methods have weaker generalization. To overcome this challenge, the other line of works, named as \emph{keypoint regression}, only utilize network for keypoint detection and regression. Then, following the conventional pipeline, the pose is then calculated based on the keypoint matching, leaving the nonlinear rotation to geometric estimator \cite{tekin2018real, peng2019pvnet, zakharov2019dpod}. Obviously, this line of works loses the end-to-end training of the network.

At the perspective of sensor modalities, with the progress in quality and lower cost, RGB-D sensor becomes a new preference for object pose estimation. With the aid of depth, the keypoints can be represented in 3D, thus overcoming the information degeneration in 3D-2D configuration. For 3D-3D correspondences, a closed form solution can be achieved, avoiding more degenerated cases \cite{he2020pvn3d, besl1992method}. In addition, depth also provide important discriminative cues for occlusion and cluttered scenes, which is common in robotic objects manipulation \cite{mahler2017dex}.

In this paper, we set to build an RGB-D object pose estimator, which integrates the end-to-end learning in \emph{direct regression}, into the \emph{keypoint regression}, so that geometric estimator is employed without losing the end-to-end training. The main ideas of these different lines of works are illustrated in Fig. \ref{fig:method}. Moreover, we introduce the differentiable outliers elimination into the geometric estimator, so that initial pose solution can be estimated more robustly. Then, the pose is further improved via the iterative refinement \cite{wang2019densefusion}. Note that this pipeline imitates the conventional pipeline more tightly, which robustly estimates the initial pose using robust solver like RANSAC \cite{fischler1981random}, and refines the pose using nonlinear optimization \cite{wang2019densefusion}. Our psychology is to leave the feature construction to learning as it is conventionally achieved by empirical design, but keep geometry to calculus and optimization as it is already guaranteed by theory.

In summary, this work has the following contributions:
\begin{itemize}
    \item A differentiable outliers elimination mechanism is designed by softly aggregating the keypoints positions and candidate poses from minimal solvers bank, so that the robustness of pose estimation is improved.  	
    \item An end-to-end object pose estimation network is proposed by differentiating the geometric estimator, so that gradients can be back-propagated from final pose loss to the keypoint prediction.
    \item The performance of the proposed method is evaluated on three large public benchmark datasets, YCB-Video, LineMOD and Occlusion LineMOD, showing state-of-the-art performance with only pose annotation.
\end{itemize}

\section{Related work}
\label{sec:Related Work}

\subsection{Template matching methods}
While object CAD model is available in this task, some traditional methods select the most appropriate feature embedding of object model compared with the scene data and calculate the pose transformation. PPF \cite{drost2010model} employs point cloud and proposes a characteristic point pair feature. During inference phase, the ppf features of scene points are matched with the ppf feature of model points and cast votes to pose. \cite{hinterstoisser2016going} proposes novel sampling and voting strategies to avoid sensor noise and background clutter. Hinterstoisser \cite{hinterstoisser2011multimodal, hinterstoisser2012model} calculates contour gradient vectors from color image and surface normal vectors from depth image respectively to compose multimodal features for template matching. AAE \cite{sundermeyer2018implicit} proposes an augmented auto-encoder to encode only pose information of models from various views, then compares the similarity with real image embedding. CosyPose \cite{labbe2020cosypose} carries out the matching and joint optimization of objects in the whole scene based on the observation of multiple views. Template matching based methods can easily estimate the general result with exact CAD model of the object. But a great quantity of matching is time-consuming, and the pose obtained is the discrete result, which leads to less accuracy.

\subsection{Learning based regression methods}
Due to the strong fitting ability of CNN, there are also some methods which encode features from RGB or RGB-D data and directly classify or regress pose. SSD-6D \cite{kehl2017ssd} extends SSD \cite{liu2016ssd} framework and classifies decomposed pose in anchor-based way. PoseCNN \cite{xiang2017posecnn} votes for center translation and regresses rotation respectively from RGB image. MCN \cite{li2018unified} decouples translation and rotation, then divides them into several bins for classification. DenseFusion \cite{wang2019densefusion} proposes a dense fusion strategy which fuses color embedding and geometric embedding in point-wise way, then regresses pose directly. The question of classification is still the imprecision caused by discretization. As for direct regression, learning ability of the network is limited because of the nonlinearity of rotation space. Moreover, there is no special treatment for occlusion in this way.

\subsection{2D-3D correspondence methods}
Inspired by keypoint detection based on RGB image, many methods convert this issue to 2D projection of keypoint location, and then calculate pose by PnP algorithm to obtain 2D-3D correspondence. BB8 \cite{rad2017bb8} employs CNN to detect eight corners of 3D bounding box. YOLO-6D \cite{tekin2018real} and 3D-SSD \cite{luo20203d} extend YOLO and SSD respectively to predict 3D bounding box corners. However, the corners of bounding box are not on the surface of the object, which leads to large localization errors. Therefore, PVNet \cite{peng2019pvnet} employs the farthest point sampling (FPS) algorithm to select more representative keypoints. Besides, PVNet predicts the direction vector from each pixel pointing to the projection point and employs RANSAC voting strategy to locate the projection points, which can acquire more robust results. \cite{yu20206dof} further proposes DPVLoss to constrain the distance between point and vector, which makes vector prediction more robust. More recent methods develop dense prediction of 2D-3D correspondence and estimate pose by PnP-RANSAC. Pix2Pose \cite{park2019pix2pose} generates 3D coordinate mapping image from 2D pixels of input image based on GAN training. DPOD \cite{zakharov2019dpod} predicts dense 2D-3D mapping between RGB image and model with UV texture map projection. EPOS \cite{hodan2020epos} predicts 2D-3D correspondence based on fragments and estimates pose based on a variant of PnP-RANSAC. These methods based on RGB image need two-stage inference, and rely on the quality of point prediction. DSAC \cite{brachmann2017dsac} proposes differentiable ransac for camera localization. \cite{hu2020single} makes efforts for combining the two stage into one for end-to-end training, but the pose is inferred by MLP, which still cannot overcome the shortcoming of direct regression lines.

\begin{figure*}[thpb]
	\centering
	\includegraphics[width=1\linewidth]{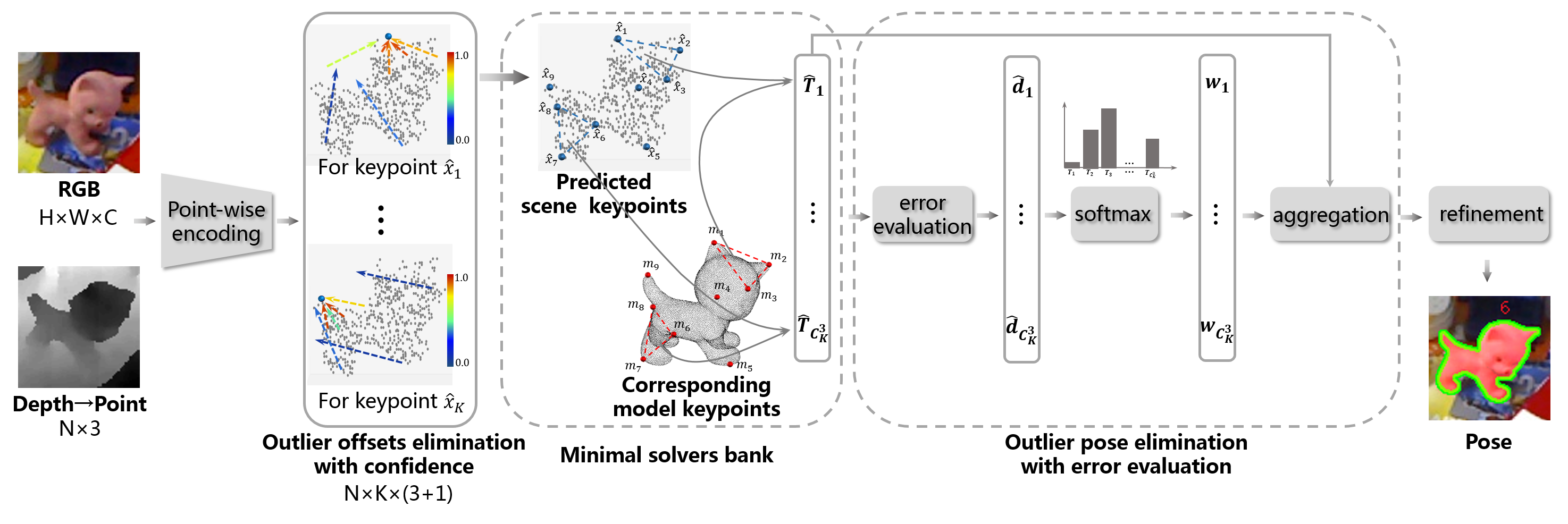}
	\caption{Overview of REDE: The point-wise point-to-keypoint offsets are predicted from the segmented RGB-D frame. We predict the corresponding confidences meanwhile for the robust aggregation of 3D keypoints. Then multiple poses are estimated by exhaustively enumerating every 3 keypoints among all candidate keypoints to build minimal solvers bank. Next, the pose is estimated robustly by weighted aggregation with differentiable outliers elimination and final refinement.}
	\label{fig:pipeline}
	\vspace{-0.2cm}
\end{figure*}

\section{End-to-end robust pose estimation}
\label{sec:Methodology}

An overview of REDE is shown in Fig. \ref{fig:pipeline}. In the first stage, the RGB-D frame is fed into the network to generate point-wise point-to-keypoint offsets and confidences. In the second stage, we calculate the 3D keypoints by aggregating the offsets. In the third stage, we estimate multiple poses and their confidences by exhaustively enumerating every 3 keypoints among all candidate keypoints to build minimal solvers bank. The confidence is a mapping of the residue between scene points and model points transformed using the corresponding estimated pose, leading to a robust aggregation to outlier pose. At last, the initial result is further refined for better accuracy.

\subsection{RGB-D Feature Encoding}

Before pose estimation, we need to segment region of interest in image and point cloud. Since segmentation is not the subject of our study, we use the existing segmentation results of \cite{xiang2017posecnn}, or ground truth masks as previous works \cite{he2020pvn3d}. Then we follow \cite{wang2019densefusion} to extract point-wise feature embedding: the RGB image segment is encoded into color embedding using PSPNet \cite{zhao2017pyramid}, and the point cloud segment is encoded into geometric embedding using PointNet \cite{qi2017pointnet}. To build the point-wise features, the geometric embedding is concatenated with its corresponding color embedding. Then the global feature vector is generated by average pooling layer and further concatenates to each point feature. After these processes, we build dense point-wise feature embedding containing both color and geometric information, as well as local and global information.

\subsection{Robust 3D Keypoint Prediction}

We take 3D keypoint as a mediator to estimate pose. As \cite{peng2019pvnet}, we employ the farthest point sampling (FPS) algorithm to sample 3D keypoints $\{m_k\}_{k=1}^K$ from the CAD model of each object. Then it is required to locate corresponding 3D keypoints in the current RGB-D image, which in previous works is achieved by regressing the relative direction from each point to the keypoint, followed by an aggregation of all estimations. This method is considered to be more stable, and learn the spatial structure characteristics of point cloud more effectively. However, this process is not differentiable, thus \cite{peng2019pvnet} has no back-propagation after keypoint regression.

\textbf{Offset regression:} To solve this problem, thanks to the RGB-D information, we predict the offset from each point to the keypoint, which is also followed an aggregation of all estimations. In this way, the aggregation is simply a closed form of averaging, thus the keypoint regression becomes differentiable. In detail, for the keypoint of scene $\{x_k\}_{k=1}^K$, the offset from input scene point $\{s_i\}_{i=1}^N$ should be:
\begin{equation}
v_{k,i} = x_k-s_i
%\label{eq:eq1}
\end{equation}

We utilize the smooth L1 error in $x$, $y$ and $z$ directions as the loss term. The loss term for offset vector prediction is defined as the following equations:
\begin{equation}
\mathcal{L}oss_{vec} = \sum_{k=1}^K \sum_{i=1}^NL(\Delta v_{k,i}|_x)+L(\Delta v_{k,i}|_y)+L(\Delta v_{k,i}|_z)
\label{lossvec}
\end{equation}
\begin{equation}
\Delta v_{k,i} = \hat v_{k,i}-v_{k,i}
\end{equation}
\begin{equation}
L(x) =
\left\{
\begin{array}{lr}
0.5x^2, &  if (|x|<1) \\
|x|-0.5, &   otherwise
\end{array}
\right.
\end{equation}
where $\hat v_{k,i}$ is the predicted offset, $\cdot |_x$, $\cdot |_y$ and $\cdot |_z$ are the $x$, $y$ and $z$ component of $\cdot$.

\textbf{Outlier offset elimination:} Since predicted offsets with large errors are inevitable, which are outliers in estimation, the simple averaging for offset prediction is not robust. Therefore, we design a differentiable outliers elimination method for robust keypoint prediction. In addition to the offset, we also regress confidence for each offset, denoted as $c_{k,i}$. With the confidence $c_{k,i}$, the predicted offset $\hat v_{k,i}$ is softly aggregated to get a position of the keypoint $\hat x_k$ as
\begin{equation}
\hat x_k = \sum_{i=1}^N c_{k,i} (s_i+\hat v_{k,i})
\end{equation}
In this way, confidence can be learned using pose loss.

\subsection{Robust Differentiable Pose Estimator}

Given the predicted 3D keypoints $\{\hat x_k\}_{k=1}^K$ , we can solve the object pose based on the keypoints $\{m_k\}_{k=1}^K$ in the object coordinate. The solver is built upon the optimization of the relative distance between the predicted keypoints, and the transformed model keypoints using the object pose. It can be written as
\begin{equation}
\hat R,\hat t = \arg\min_{R,t}{\sum_{k=1}^K||(R\cdot m_k+t)-\hat x_k||^2}
\label{con:leastsquare}
\end{equation}
where $R$ and $t$ form the 6D object pose. Fortunately, this optimization problem can be solved by SVD in a closed form. Therefore, it can be easily embedded in our network without losing end-to-end training.

\begin{figure}[t]
	\centering
	\includegraphics[width=1\linewidth]{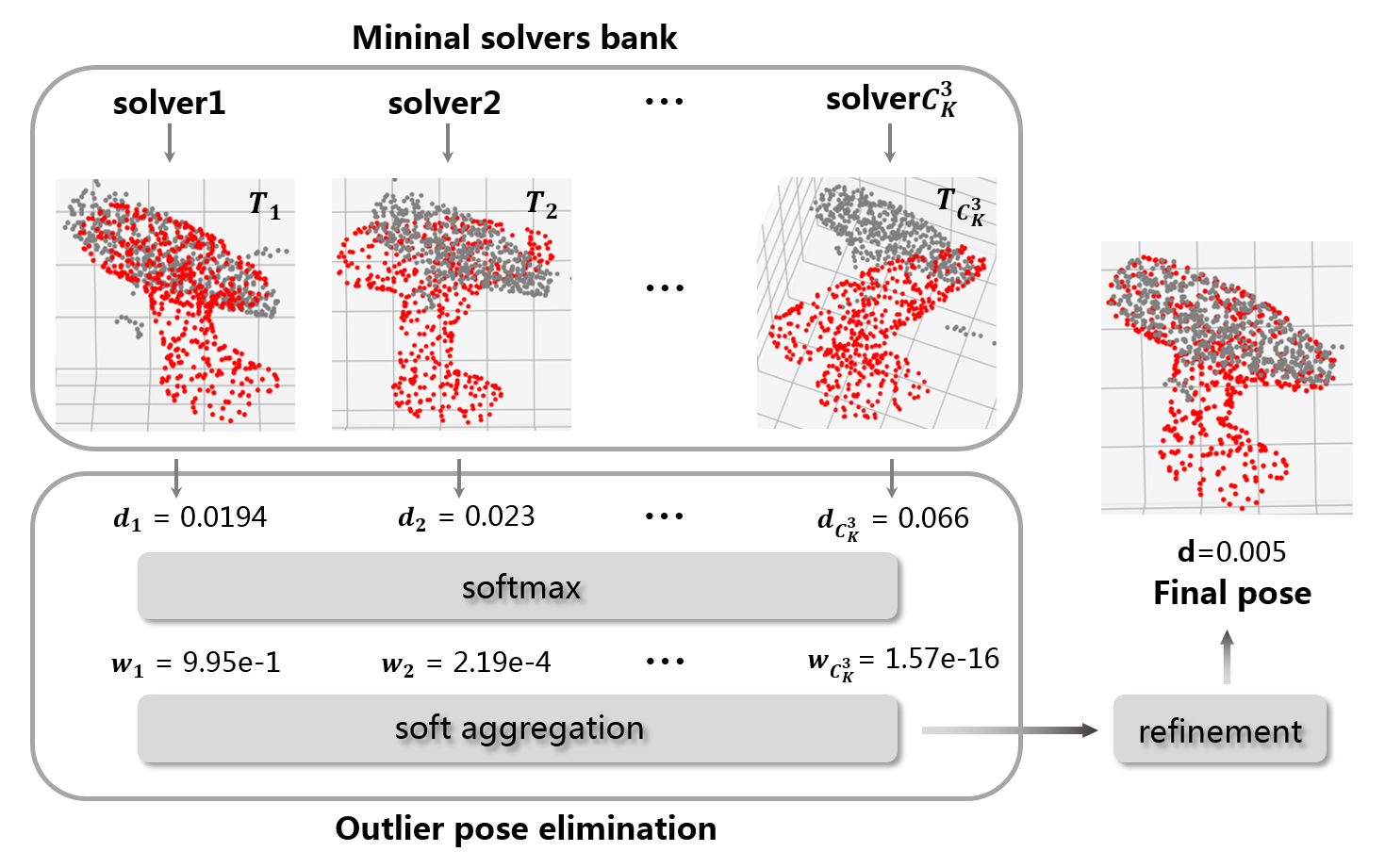}
	\caption{A case of our pose estimator using minimal solvers bank and differentiable outliers elimination mechanism, in which "driller" is under severe occlusion. For each candidate pose from minimal solvers bank, the residue between scene points and model points transformed using the corresponding estimated pose is calculated and mapped to confidence. Outlier pose derived using incorrect keypoint is assigned to low confidence, leading to a robust aggregation.}
	\label{fig:doe}
	\vspace{-0.2cm}
\end{figure}

\textbf{Minimal solvers bank:} Since the RGB-D data can only be captured in one view, there can be some unobservable keypoints due to occlusion. Therefore, it is highly possible that the network can not precisely predict them. As a result, the potential outliers may degenerate the performance seriously. A simple idea is to follow the outlier elimination in keypoint prediction that we can assign a weight for each keypoint and employ the weighted least square algorithm. But we find that the performance becomes worse, probably because the network is prone to give up many keypoints. DSAC \cite{brachmann2017dsac} proposes differentiable ransac in the field of camera localization. But it is used for dense points, which has to adopt stochastic sampling strategy.

To make the estimator more robust, we propose minimal solvers bank. For every 3 keypoints, we can solve a pose as a candidate using the minimal version of (\ref{con:leastsquare}). Thanks to the small number of keypoints, we can permutate all possibilities to generate a bank of $C_K^3$ minimal solvers. This module can be evaluated efficiently based on GPU and closed form.

\textbf{Outlier pose elimination:} Given the resultant set of candidate pose $\{\hat T_i\}_{i=1}^{C_K^3}$, which can be split into rotation $\{\hat R_i\}_{i=1}^{C_K^3}$ and translation $\{\hat t_i\}_{i=1}^{C_K^3}$, we need to aggregate them to generate a robust estimation. We still use confidence to softly average the candidates. However, different from the outlier offset elimination, we first calculate the error distance between scene points $\{s_j\}_{j=1}^N$ and their nearest neighbor among the transformed model points $\{p_j\}_{j=1}^N$ using the corresponding candidate pose, say $\hat T_i$ 
\begin{equation}
\hat d_i = {\sum_{j=1}^N||({\hat R_i}\cdot p_{N(s_j)}+{\hat t_i})-s_j||}
\end{equation}
where $p_{N(s_j)}$ is the nearest neighbor of the scene point $s_j$ among the transformed model points. Then the confidence $w_i$ is predicted as a monotonic mapping of the error distance. Repeating the process for all the candidate poses, we have a confidence vector for all candidates in $\{\hat T_i\}_{i=1}^{C_K^3}$.

To normalize the weights, we apply softmax to each error $d_i$ to derive the final weights $\{ w_i\}_{i=1}^{C_K^3}$ as
\begin{equation}
w_i = \frac {e^{-\frac{\hat d_i}{\lambda}}}{\sum_{j=1}^{C_K^3}{e^{-\frac{\hat d_j}{\lambda}}}}
\end{equation}
where $\lambda$ is a temperature coefficient. With the weights, we can softly aggregate the translation by weighted averaging
\begin{equation}
\hat t = \sum_{i=1}^{C_K^3} w_{i}\cdot \hat t_{i}
\end{equation}

For rotation, the aggregation is slightly different due to the nonlinear structure of the rotation space. We first transform the rotation matrix $\{\hat R_i\}_{i=1}^{C_K^3}$ into quaternion $\{\hat q_i\}_{i=1}^{C_K^3}$ for aggregation since averaging the rotation matrices can break the $SO(3)$ constraints. After that, the quaternions are aggregated and normalized to get the unit vector, which is depicted as:

\begin{equation}
q' = \sum_{i=1}^{C_K^3} w_{i}\cdot \hat q_{i}, \quad
\hat q = \frac {q'} {||q'||}
\end{equation}

\textbf{Differentiable pose estimation:} We arrive at the estimated pose $\{\hat q, \hat t\}$, which is robust to the outliers caused by incorrect offset regression. Since the whole robust pose estimator is differentiable, we can train the keypoint prediction network with error signal from not only the keypoint ground truth, but also the pose ground truth. The latter loss term is defined as 
\begin{equation}
\mathcal{L}oss_{pose} = ||\hat t-t||_2+\alpha ||\hat R\cdot R^T-I||_F
\end{equation}
where $\alpha$ is a balancing parameter, $R$ and $t$ are the ground truth pose, $\hat R$ is coverted from $\hat q$.

In total, the network is trained by a joint loss 
\begin{equation}
\mathcal{L}oss = \mathcal{L}oss_{vec} + \beta \cdot \mathcal{L}oss_{pose}
\end{equation}
where $\beta$ is a trade-off parameter between the two terms. Finally, following the conventional pipeline that nonlinearly refines the initial value derived from the robust estimator, we also further iteratively optimize the pose via iterative refinement in \cite{wang2019densefusion}. A heavily occluded case of our pose estimator using minimal solvers bank and differentiable outliers elimination mechanism is illustrated in Fig. \ref{fig:doe}.

\begin{table*}[htbp]
	\centering
	\caption{ADD-S performance on YCB-Video dataset.}
	\label{table:ycb}
	\newcommand{\tabincell}[2]{\begin{tabular}{@{}#1@{}}#2\end{tabular}}
	%\begin{tabular}{c|c|c|c|c|c|c|c|c|c|c|c|c|c|c|c|c}
	\begin{tabular}{m{0.7cm}<{\centering}|m{0.6cm}<{\centering}|m{0.6cm}<{\centering}|m{0.6cm}<{\centering}|m{0.6cm}<{\centering}|m{0.6cm}<{\centering}|m{0.6cm}<{\centering}|m{0.6cm}<{\centering}|m{0.6cm}<{\centering}|m{0.6cm}<{\centering}|m{0.6cm}<{\centering}|m{0.6cm}<{\centering}|m{0.6cm}<{\centering}|m{0.6cm}<{\centering}|m{0.6cm}<{\centering}|m{0.6cm}<{\centering}|m{0.6cm}<{\centering}}
		\hline
		& \multicolumn{8}{c}{with GT mask} & \multicolumn{8}{|c}{with PoseCNN mask} \\
		\hline
		& \multicolumn{2}{c}{\tabincell{c}{DenseFusion\\\cite{wang2019densefusion}}} & \multicolumn{2}{|c}{\tabincell{c}{P$^2$GNet\\\cite{yu2019p}}} &
		\multicolumn{2}{|c}{\tabincell{c}{PVN3D\\(w/o \\semantics)\\\cite{he2020pvn3d}}} & \multicolumn{2}{|c}{REDE} & \multicolumn{2}{|c}{\tabincell{c}{PoseCNN\\\cite{xiang2017posecnn}}} & \multicolumn{2}{|c}{\tabincell{c}{DenseFusion\\\cite{wang2019densefusion}}} & \multicolumn{2}{|c}{\tabincell{c}{Tian et al. \\\cite{tian2020robust}}} &
		\multicolumn{2}{|c}{REDE} \\
		\hline
		refine & \multicolumn{2}{c}{$\surd$} & \multicolumn{2}{|c}{$\surd$} & \multicolumn{2}{|c}{} & \multicolumn{2}{|c}{$\surd$} & \multicolumn{2}{|c}{$\surd$} & \multicolumn{2}{|c}{$\surd$} & \multicolumn{2}{|c}{} & \multicolumn{2}{|c}{$\surd$} \\
		\hline
		metric & AUC & \textless 2cm & AUC & \textless 2cm & AUC & \textless 2cm & AUC & \textless 2cm & AUC & \textless 2cm & AUC & \textless 2cm & AUC & \textless 2cm & AUC & \textless 2cm \\
		
		\hline
		002 & 96.2 & 100.0 & - & - & - & - & 95.4 & 100.0 & 95.8 & 100.0 & 96.4 & 100.0 & 93.9 & - & 95.1 & 100.0\\
		003 & 95.3 & 100.0 & - & - & - & - & 96.4 & 100.0 & 92.7 & 91.6 & 95.5 & 99.5 & 92.9 & - & 96.3 & 99.7\\
		004 & 97.9 & 100.0 & - & - & - & - & 98.0 & 100.0 & 98.2 & 100.0 & 97.5 & 100.0 & 95.4 & - & 97.4 & 100.0\\
		005 & 94.3 & 96.9 & - & - & - & - & 96.2 & 98.4 & 94.5 & 96.9 & 94.6 & 96.9 & 93.3 & - & 96.9 & 100.0\\
		006 & 97.7 & 100.0 & - & - & - & - & 98.0 & 100.0 & 98.6 & 100.0 & 97.2 & 100.0 & 95.4 & - & 96.7 & 100.0\\
		007 & 96.7 & 100.0 & - & - & - & - & 96.9 & 100.0 & 97.1 & 100.0 & 96.6 & 100.0 & 94.9 & - & 96.6 & 100.0\\
		008 & 97.3 & 100.0 & - & - & - & - & 97.8 & 100.0 & 97.9 & 100.0 & 96.5 & 100.0 & 94.0 & - & 96.4 & 100.0\\
		009 & 98.4 & 100.0 & - & - & - & - & 98.8  & 100.0 & 98.8 & 100.0 & 98.1 & 100.0 & 97.6 & - & 97.8 & 100.0\\
		010 & 90.2 & 92.3 & - & - & - & - & 92.1 & 94.5 & 92.7 & 93.6 & 91.3 & 93.1 & 90.6 & - & 92.0 & 94.2\\
		011 & 96.2 & 99.7 & - & - & - & - & 97.7 & 100.0 & 97.1 & 99.7 & 96.6 & 100.0 & 91.7 & - & 97.0 & 99.7\\
		019 & 97.5 & 100.0 & - & - & - & - & 98.1 & 100.0 & 97.8 & 100.0 & 97.1 & 100.0 & 93.1 & - & 97.5 & 100.0\\
		021 & 96.4 & 100.0 & - & - & - & - & 96.7 & 100.0 & 96.9 & 99.4 & 95.8 & 100.0 & 93.4 & - & 94.2 & 100.0\\
		024 & 88.9 & 87.4 & - & - & - & - & 96.6 & 99.5 & 81.0 & 54.9 & 88.2 & 98.8 & 92.9 & - & 96.7 & 99.3\\
		025 & 97.0 & 100.0 & - & - & - & - & 97.5 & 100.0 & 95.0 & 99.8 & 97.1 & 100.0 & 96.1 & - & 97.0 & 100.0\\
		035 & 97.1 & 100.0 & - & - & - & - & 97.9 & 100.0 & 98.2 & 99.6 & 96.0 & 98.7 & 93.3 & - & 97.0 & 99.6\\
		036 & 94.1 & 100.0 & - & - & - & - & 93.8 & 100.0 & 87.6 & 80.2 & 89.7 & 94.6 & 87.6 & - & 91.0 & 98.3\\
		037 & 93.2 & 100.0 & - & - & - & - & 93.5 & 96.7 & 91.7 & 95.6 & 95.2 & 100.0 & 95.7 & - & 94.5 & 100.0\\
		040 & 97.5 & 100.0 & - & - & - & - & 98.1 & 100.0 & 97.2 & 99.7 & 97.5 & 100.0 & 95.6 & - & 97.8 & 100.0\\
		051 & 89.7 & 98.0 & - & - & - & - & 96.9 & 100.0 & 75.2 & 74.9 & 72.9 & 79.2 & 75.4 & - & 77.3 & 80.7\\
		052 & 77.4 & 80.5 & - & - & - & - & 96.2 & 99.9 & 64.4 & 48.8 & 69.8 & 76.3 & 73.0 & - & 85.9 & 82.0\\
		061 & 91.5 & 100.0 & - & - & - & - & 95.5 & 100.0 & 97.2 & 100.0 & 92.5 & 100.0 & 94.2 & - & 94.6 & 100.0\\
		\hline
		MEAN  & 94.2 & 97.8 & 94.2 & 97.8 & 94.8 & - & \textbf{96.6} & \textbf{99.5} & 93.0 & 93.2 & 93.1 & 96.8 & 91.8 & - & \textbf{94.5} & \textbf{97.8}\\
		\hline
	\end{tabular}
	\vspace{-0.2cm}
\end{table*}

\section{Experiments}

To validate the proposed method, we report the performance by comparing it with the state-of-the-art methods on three datasets, YCB-Video dataset  \cite{xiang2017posecnn}, LineMOD dataset \cite{hinterstoisser2011multimodal} and Occlusion LineMOD dataset \cite{brachmann2014learning}.

\subsection{Datasets}

YCB-Video dataset contains 21 objects with various textrues from YCB objects. There are 92 RGB-D videos in which 80 videos are used for training and 2949 keyframes from the rest 12 videos are used for testing. Besides, 80000 synthetic images are released for training. There are many scenes of stacking objects with partial occlusion.

LineMOD dataset is a widely used benchmark for object 6D pose estimation task. This dataset contains 13 objects totally. We follow prior learning-based
works \cite{wang2019densefusion,he2020pvn3d} to split training and testing data. There are about 180 training images and 1000 testing images for each object. 10000 images using the “Cut and Paste” strategy are further synthesized for training as \cite{peng2019pvnet}.

Occlusion LineMOD dataset further annotates data with serious occlusion in LineMOD dataset. It contains 8 objects and 1214 images. All images are used for evaluation with model trained on LineMOD dataset. The main challenge on this dataset is severe occlusion, especially for small target.

\subsection{Metrics}

The most commonly used metrics for object pose estimation are ADD \cite{hinterstoisser2012model} and ADD-S \cite{xiang2017posecnn}. ADD metric is defined as the average Euclidean distance between model points transformed with the predicted and the ground truth pose respectively:

\begin{equation}
ADD = \frac 1 N\sum_{i=1}^N||(R\cdot p_i+t)-(\hat R\cdot p_i+\hat t)||
\end{equation}
where $N$ is the number of model points $\{p_i\}_{i=1}^N$, $R$ and $t$ are the rotation and translation of ground truth pose, $\hat R$ and $\hat t$ are the rotation and translation of predicted pose.

ADD-S metric is designed for symmetric object and calculates the average distance with the closest point:

\begin{equation}
{ADD\text -S} = \frac 1 N\sum_{i=1}^N\min_{j\in [1,N]}||(R\cdot p_j+t)-(\hat R\cdot p_i+\hat t)||
\end{equation}

For YCB-Video dataset, we report the results of two metrics with ADD-S, the same as \cite{wang2019densefusion}. The first is the correct percentage in all data which ADD-S is smaller than 2cm. The second is the area under the ADD-S curve (AUC), which is obtained by varying the distance threshold in evaluation.

For LineMOD dataset and Occlusion LineMOD dataset, we use ADD(-S) metric following prior works \cite{hinterstoisser2012model}. For non-symmetric objects, we use ADD metric. For symmetric objects (eggbox and glue in two datasets), due to the ambiguity of pose, we use ADD-S metric.
We regard the evaluation result as accurate if ADD(-S) is less than 10\% of the object model’s diameter.

\begin{table*}
	\centering
	\caption{ADD(-S) performance on LineMOD dataset.}
	\label{table:linemod}
	\newcommand{\tabincell}[2]{\begin{tabular}{@{}#1@{}}#2\end{tabular}}
	\begin{threeparttable}
		%\begin{tabular}{c|c|c|c|c|c|c|c|c}
		\begin{tabular}{m{1.5cm}<{\centering}|m{1.5cm}<{\centering}|m{1.5cm}<{\centering}|m{1.5cm}<{\centering}|m{1.5cm}<{\centering}|m{1.5cm}<{\centering}|m{1.5cm}<{\centering}|m{1.5cm}<{\centering}|m{1.5cm}<{\centering}}
			\hline
			& \multicolumn{3}{c}{RGB} & \multicolumn{5}{|c}{RGB-D\tnote{1}} \\
			\hline
			%& PVNet \cite{peng2019pvnet} & DPOD \cite{zakharov2019dpod} & DPVL \cite{yu20206dof} & PointFusion \cite{xu2018pointfusion} & DenseFusion \cite{wang2019densefusion} & P$^2$GNet \cite{yu2019p} & Tian \cite{tian2020robust} & REDE\\
			& \tabincell{c}{PVNet \\\cite{peng2019pvnet}} & \tabincell{c}{DPOD \\\cite{zakharov2019dpod}} & \tabincell{c}{DPVL \\\cite{yu20206dof}} & \tabincell{c}{PointFusion \\\cite{xu2018pointfusion}} & \tabincell{c}{DenseFusion \\\cite{wang2019densefusion}} & \tabincell{c}{P$^2$GNet \\\cite{yu2019p}} & \tabincell{c}{Tian et al. \\\cite{tian2020robust}} & REDE\\
			\hline
			ape & 43.6 & 87.7 & 69.1 & 70.4 & 92.3 & 92.9 & 85.0 & 95.6\\
			benchvise & 99.9 & 98.5 & 100.0 & 80.7 & 93.2 & 98.2 & 95.5 & 99.4\\
			cam & 86.9 & 96.1 & 94.1 & 60.8 & 94.4 & 97.0 & 91.3 & 99.6\\
			can & 95.5 & 99.7 & 98.5 & 61.1 & 93.1 & 97.4 & 95.2 & 99.5\\
			cat & 79.3 & 94.7 & 83.1 & 79.1 & 96.5 & 98.1 & 93.6 & 99.5\\
			driller & 96.4 & 98.8 & 99.0 & 47.3 & 87.0 & 97.0 & 82.6 & 99.3\\
			duck & 52.6 & 86.3 & 63.5 & 63.0 & 92.3 & 95.2 & 88.1 & 97.0\\
			eggbox & 99.2 & 99.9 & 100.0 & 99.9 & 99.8 & 100.0 & 99.9 & 100.0 \\
			glue & 95.7 & 96.8 & 98.0 & 99.3 & 100.0 & 100.0 & 99.6 & 99.9 \\
			holepuncher & 81.9 & 87.7 & 88.2 & 71.8 & 86.9 & 97.9 & 92.6 & 98.6\\
			iron & 98.9 & 100.0 & 99.9 & 83.2 & 97.0 & 98.2 & 95.9 & 99.3\\
			lamp & 99.3 & 96.8 & 99.8 & 62.3 & 95.3 & 97.7 & 94.4 & 99.3\\
			phone & 92.4 & 94.7 & 96.4 & 78.8 & 92.8 & 96.7 & 93.6 & 99.3\\
			\hline
			average & 86.3 & 95.2 & 91.5 & 73.3 & 94.3 & 97.4 & 92.9 & \textbf{98.9} \\
			\hline
		\end{tabular}
		\begin{tablenotes}
			\footnotesize
			\item[1] All the RGB-D based methods listed here use PoseCNN mask.
			%		\item[2] bb
		\end{tablenotes}
	\end{threeparttable}
	\vspace{-0.2cm}
\end{table*}

\begin{table}
	\centering
	\caption{ADD(-S) performance on Occlusion LineMOD dataset.}
	\label{table:occ}
	\newcommand{\tabincell}[2]{\begin{tabular}{@{}#1@{}}#2\end{tabular}}
	%\begin{tabular}{c|c|c|c|c|c}
	%\begin{tabular}{m{1.2cm}<{\centering}|m{0.9cm}<{\centering}|m{0.9cm}<{\centering}|m{0.9cm}<{\centering}|m{0.9cm}<{\centering}|m{0.9cm}<{\centering}}
	\begin{threeparttable}
		\begin{tabular}{m{1.2cm}<{\centering}|m{0.8cm}<{\centering}|m{0.8cm}<{\centering}|m{0.8cm}<{\centering}|m{0.8cm}<{\centering}|m{0.8cm}<{\centering}}
			\hline
			%& \tabincell{c}{PVNet \\\cite{peng2019pvnet}} & \tabincell{c}{Pix2Pose \\\cite{park2019pix2pose}} & \tabincell{c}{DPOD \\\cite{zakharov2019dpod}} & \tabincell{c}{DPVL \\\cite{yu20206dof}} & Ours\\
			& PVNet \cite{peng2019pvnet} & DPOD \cite{zakharov2019dpod} & DPVL \cite{yu20206dof} & PVN3D \cite{he2020pvn3d} & REDE\tnote{1}\\
			\hline
			ape & 15.8 & - & 19.2 & 33.9 & 53.1\\
			can & 63.3 & - & 69.8 & 88.6 & 88.5\\
			cat & 16.7 & - & 21.1 & 39.1 & 35.9\\
			driller & 65.7 & - & 71.6 & 78.4 & 77.8\\
			duck & 25.2 & - & 34.3 & 41.9 & 46.2\\
			eggbox & 50.2 & - & 47.3 & 80.9 & 71.8\\
			glue & 49.6 & - & 39.7 & 68.1 & 75.0\\
			holepuncher & 39.7 & - & 45.3 & 74.7 & 75.5\\
			\hline
			average & 40.8 & 47.3 & 43.5 & 63.2 & \textbf{65.4} \\
			\hline
		\end{tabular}
		\begin{tablenotes}
			\footnotesize
			\item[1] With the same mask as PVNet \cite{peng2019pvnet}.
			%		\item[2] bb
		\end{tablenotes}
	\end{threeparttable}
	\vspace{-0.2cm}
\end{table}

\begin{table}
	\centering
	\caption{Ablation studies for loss.}
	\label{table:ablation1}
	\begin{threeparttable}
		%	\begin{tabular}{c|c}
		\begin{tabular}{m{2cm}<{\centering}m{2cm}<{\centering}|m{2cm}<{\centering}}
			\hline
			offset loss & pose loss & ADD(-S)\textless 0.1d \\ 
			\hline
			$\surd$ &  & 59.6 \\
			$\surd$ & $\surd$ & 65.4 \\
			\hline
		\end{tabular}
		\begin{tablenotes}
			\footnotesize
			\item[1] With the same mask as PVNet \cite{peng2019pvnet} on Occlusion LineMOD dataset.
		\end{tablenotes}
	\end{threeparttable}
	\vspace{-0.2cm}
\end{table}

\begin{table}
	\centering
	\caption{Ablation studies for differentiable outliers elimination.}
	\label{table:ablation2}
	\begin{threeparttable}
		%	\begin{tabular}{cc|cc}
		\begin{tabular}{m{1.5cm}<{\centering}m{1.5cm}<{\centering}|m{1.5cm}<{\centering}m{1.5cm}<{\centering}}
			%\begin{tabular}{m{3.5cm}<{\centering}|m{1cm}<{\centering}|m{1cm}<{\centering}|m{1cm}<{\centering}}
			\hline
			DOE for keypoint & DOE for pose & ADD-S AUC & ADD(-S) AUC \\ 
			\hline
			&  & 92.3 & 85.4 \\
			$\surd$ &  & 92.7 & 86.6 \\
			$\surd$ & $\surd$ & 94.3 & 89.5 \\
			\hline
		\end{tabular}
		\begin{tablenotes}
			\footnotesize
			\item[1] With PoseCNN mask on YCB-Video dataset.
			%		\item[2] bb
		\end{tablenotes}
	\end{threeparttable}
	\vspace{-0.2cm}
\end{table}

\subsection{Analysis}

We explore the effectiveness of our differentiable outliers elimination mechanism for keypoint location and pose estimation in this part.

\begin{figure}[t]
	\centering
	\includegraphics[width=1\linewidth]{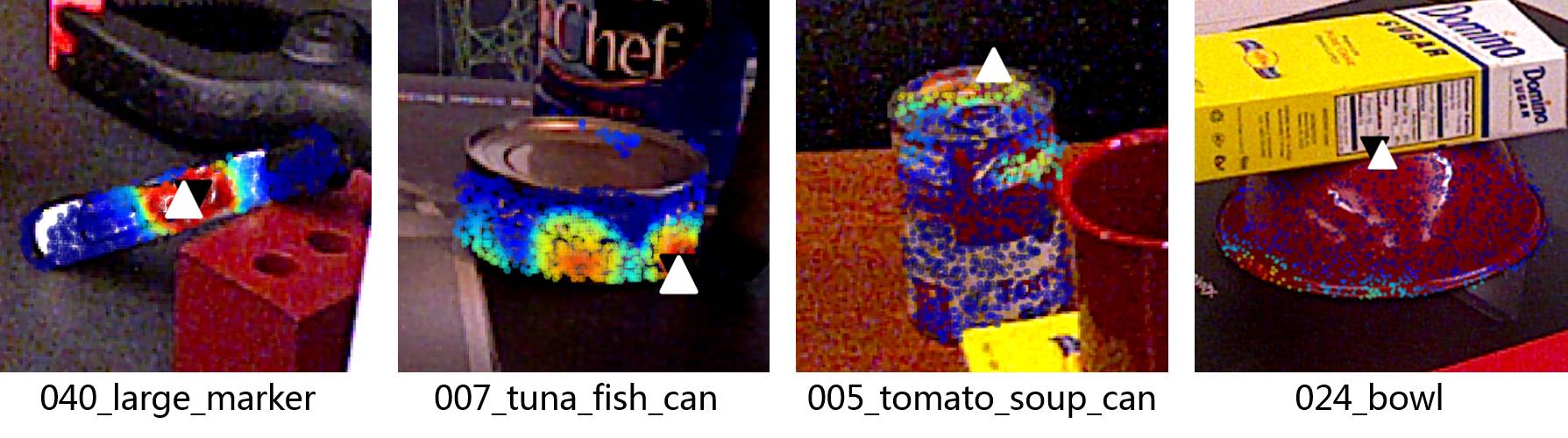}
	\caption{Visualization for confidence of offsets. Red color indicates high confidence and blue indicates low confidence. The ground truth keypoint and predicted keypoint are marked with white and black triangle dots respectively.}
	\label{fig:confidence}
	\vspace{-0.2cm}
\end{figure}

\textbf{Confidence visualization:} The confidences of offsets are visualized to observe the effect of differentiable outliers elimination mechanism in keypoint location. As shown in Fig. \ref{fig:confidence}, we project points on RGB image with varying colors to show confidences. The confidence increases as color changes from blue to red. We can see from "040\_larger\_marker" that the offset estimated by the point close to the keypoint tends to be assigned to higher confidence, which is in line with our intuition. Besides, the textured areas of "007\_tuna\_fish\_can" and "005\_tomato\_soup\_can" also have higher confidence. More interestingly, the confidences of the edge points in "024\_bowl" are higher, but the confidences of the points around the keypoint are low instead. It can be concluded that the prediction of the points with small offsets and obvious characteristics is more accurate, so the confidence is higher.

\textbf{Ablation studies:} We conduct two ablation studies to verify the effects of different losses and differentiable outliers elimination(DOE). The addition of pose loss achieves 5.8\% improvement on Occlusion LineMOD dataset (see Table \ref{table:ablation1}), which verify the importance of end-to-end training. As for our differentiable outliers elimination, table \ref{table:ablation2} summarizes the results of ablation studies on YCB-Video dataset. Here we report the results using PoseCNN mask without ICP. For ADD-S metric, Our AUC is 0.4\% higher with DOE for keypoint location and 2.0\% higher with DOE for pose estimation. For ADD(-S) metric, Our AUC is 1.2\% higher with DOE for keypoint location and 4.1\% higher with DOE for pose estimation.

\subsection{Results on Benchmark Dataset}

\begin{figure*}[t]
	\centering
	\includegraphics[width=1\linewidth]{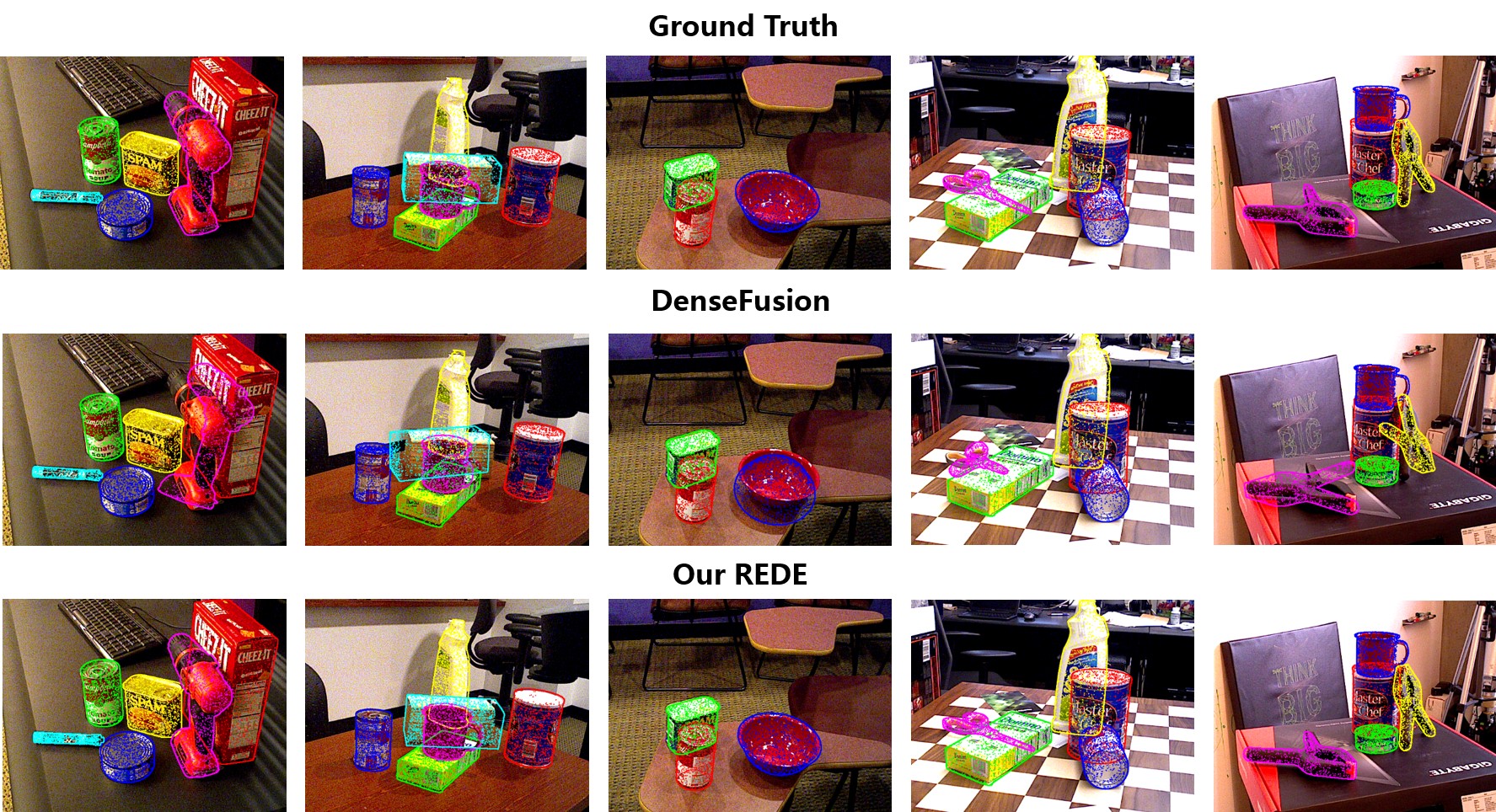}
	\caption{Some visualization results on YCB-Video dataset. The point cloud of all objects are transformed with the predicted pose and projected to the RGB image using different colors. Compared with DenseFusion, our REDE estimates the poses more accurately.
	}
	\label{fig:ycb}
	\vspace{-0.3cm}
\end{figure*}

\textbf{Evaluation on YCB-Video Dataset.} Cause the large clamp and extra large clamp in YCB-Video dataset are the same models with two different scales, the segmentation network with RGB image as the only input can not distinguish them. In order not to be affected by the ambiguous segmentation results from PoseCNN \cite{xiang2017posecnn}, we also employ ground truth mask to verify our pose estimation network. The evaluation results for this two kind of masks are both listed in Table \ref{table:ycb}. All methods listed in the table are based on RGB-D data and use only pose related supervision. REDE outperforms others on both two metrics and two kind of masks. The numerical values we report in table are the results after ICP refinement. The ADD-S AUC of our method using ground truth mask is 96.2\% without ICP and 95.6\% without refinement, which also surpasses 94.8\% in PVN3D \cite{he2020pvn3d}. Fig. \ref{fig:ycb} displays some visualization results. It can be observed that compared with DenseFusion \cite{wang2019densefusion}, our method can accurately estimate the pose of objects, especially in some hard cases with occlusion.

\begin{figure}[t]
	\centering
	\includegraphics[width=0.9\linewidth]{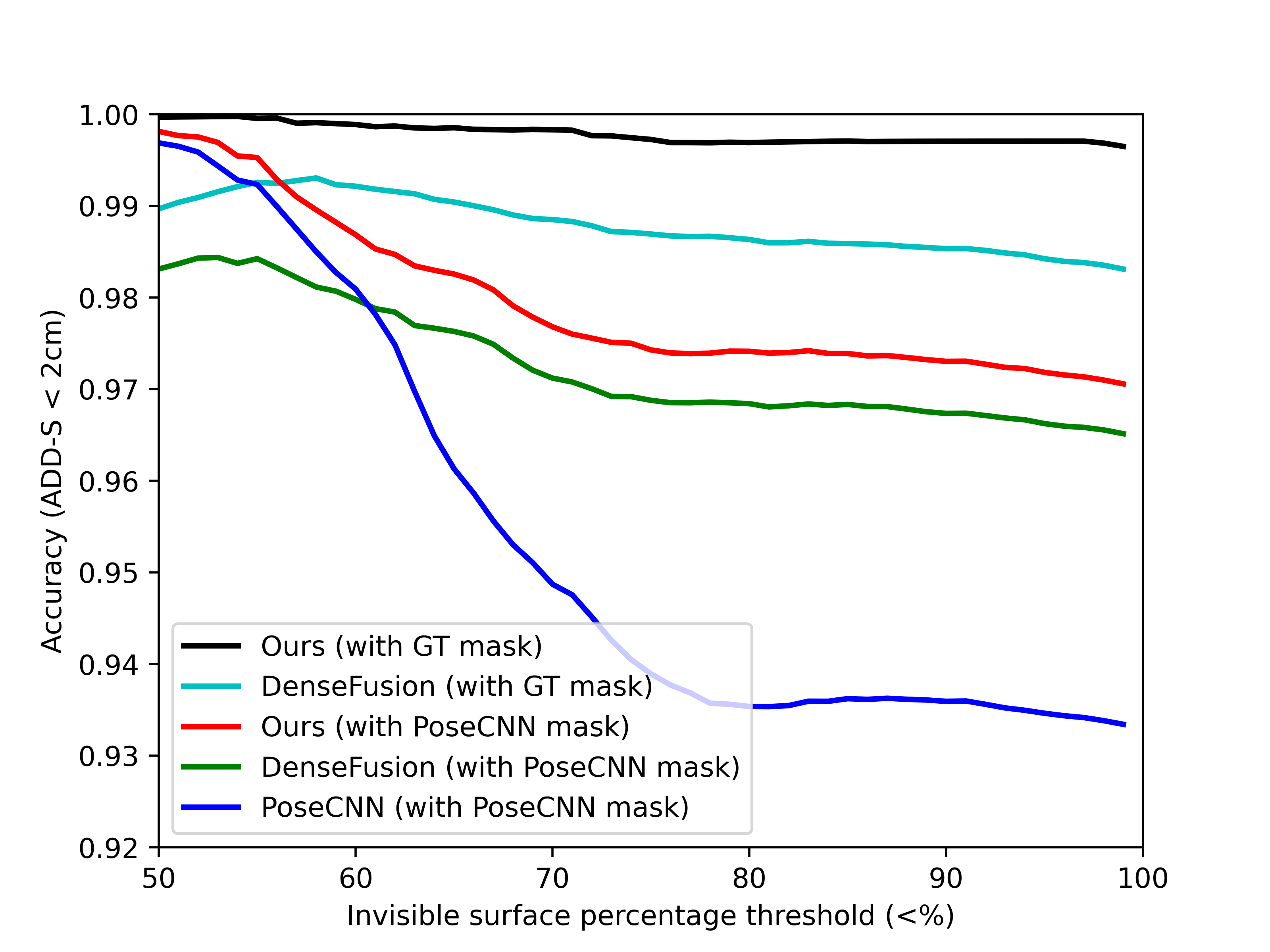}
	\caption{Performance curve for different invisible surface percentage. REDE has more robust performance under heavy occlusion especially when not affected by bad segmentation results.}
	\label{fig:curve}
	\vspace{-0.2cm}
\end{figure}

To verify our robustness towards occlusion, we also draw accuracy curve under increasing levels of occlusion on YCB-Video dataset. Following DenseFusion \cite{wang2019densefusion}, levels of occlusion are measured by calculating the invisible surface percentage of model points in the image frame. The accuracy of ADD-S smaller than 2cm curve is shown in  Fig. \ref{fig:curve}, it can be seen that our performance under occlusion is more stable, especially in the case of accurate segmentation results.

\textbf{Evaluation on LineMOD and Occlusion LineMOD Dataset.} Our quantitative evaluation results of the pose estimation experiments on the LineMOD dataset are reported in Table \ref{table:linemod}. We have to clarify that
all the RGB-D based methods listed here use the same segmentation masks released by PoseCNN. The ADD(-S) metric of our method outperforms all other approaches. Especially, the performance of prior works on the small objects such as "ape", "cat" and "duck" are poor due to the few available pixels in the image frame. But we can handle them very well which is benefit from end-to-end learning and differentible outliers elimination.

As for Occlusion LineMOD dataset with many hard cases, we use the same masks as PVNet \cite{peng2019pvnet}. We report the quantitative results in Table \ref{table:occ}, where PVN3D \cite{he2020pvn3d} is also based on RGB-D. We achieve accuracy of 65.4\% which outperforms other recent methods. This proves that our differentiable outliers elimination mechanism works well for partial occlusion, and the robustness of the estimation is improved. %Some visualization results are shown in Fig. \ref{fig:linemod}.

\subsection{Runtime}

On GTX 2080 Ti GPU, REDE takes 0.03 seconds for pose estimation and refinement. With 0.03 seconds for prior instance segmentation, the overall runtime is about 17 FPS on the LineMOD dataset, which is promising for real-time
applications.

%More details and results please refer to \cite{hua2020rede}.

\section{Conclusion}

We present REDE, an end-to-end object 6D pose robust estimation method based on keypoint regression in this paper. We integrate the end-to-end learning into keypoint regression, so as to get better supervision. In addition, we design a differentiable outliers elimination mechanism for both keypoint location and pose estimation, which solves the problem of keypoint prediction deviation probably caused by occlusion. Experiments show that REDE outperforms the state-of-the-art methods in several datasets and is especially robust for outliers.

In the future, we will consider some studies combined with other fields. First, multi-task learning will be conducted together with segmentation. Second, we will make use of the advantages of end-to-end manner for self-supervised learning. Finally, conducting transfer learning from synthetic data only is also worth to be studied further.

%\printbibliography
\bibliographystyle{IEEEtran}
\bibliography{IEEEabrv,paper}

\clearpage

\appendices   %仅一个附录时用appendix，否则\appendices
\setcounter{table}{0}   %从0开始编号，显示出来表会A1开始编号
\setcounter{figure}{0}
%定义编号格式，在数字序号前加字符“A"
\renewcommand{\thetable}{A\arabic{table}}
\renewcommand{\thefigure}{A\arabic{figure}}

\begin{table}
	\centering
	\caption{Ablation study for end-to-end training on Occlusion LineMOD dataset.}
	\label{table:e2e}
	%	\arrayrulecolor{black}
	%	\begin{tabular}{c|c|c|c|c}
	\begin{tabular}{m{1.2cm}<{\centering}|m{1.2cm}<{\centering}|m{1.2cm}<{\centering}|m{1.2cm}<{\centering}|m{1.2cm}<{\centering}}
		\hline
		& \multicolumn{3}{c}{w/o end-to-end} & \multicolumn{1}{|c}{end-to-end} \\
		\hline
		& DPVL (3D version) & PVNet (3D version) & PVNet (3D version with RANSAC) & REDE\\
		\hline
		estimator & SVD 3D-3D & SVD 3D-3D & RANSAC 3D-3D & DOE\\
		\hline
		ape & 55.7 & 51.6 & 58.3 & 53.1\\
		can & 73.2 & 75.6 & 74.9 & 88.5\\
		cat & 16.9 & 28.7 & 30.2 & 35.9\\
		driller & 68.5 & 66.9 & 73.3 & 77.8\\
		duck & 39.3 & 36.7 & 35.2 & 46.2\\
		eggbox & 39.8 & 47.1 & 46.8 & 71.8\\
		glue & 67.7 & 71.9 & 74.5 & 75.0\\
		holepuncher & 50.1 & 45.7 & 53.8 & 75.5\\
		\hline
		average & 51.1 & 52.6 & 55.5 & 65.4 \\
		\hline
	\end{tabular}
\end{table}

\begin{table}
	\centering
	\caption{Ablation studies for differentiable outliers elimination using ground truth mask.}
	\label{table:gtablation}
	\begin{tabular}{cc|cc}
		%\begin{tabular}{m{3.5cm}<{\centering}|m{1cm}<{\centering}|m{1cm}<{\centering}|m{1cm}<{\centering}}
		\hline
		DOE for keypoint & DOE for pose & ADD-S AUC & ADD(-S) AUC \\ 
		\hline
		&  & 94.9 & 89.3 \\
		$\surd$ &  & 95.2 & 90.2 \\
		$\surd$ & $\surd$ & 96.2 & 92.2 \\
		\hline
	\end{tabular}
\end{table}

\begin{table}
	\centering
	\caption{Ablation studies for differentiable outliers elimination using PoseCNN mask.}
	\label{table:pcablation}
	\begin{tabular}{cc|cc}
		%\begin{tabular}{m{3.5cm}<{\centering}|m{1cm}<{\centering}|m{1cm}<{\centering}|m{1cm}<{\centering}}
		\hline
		DOE for keypoint & DOE for pose & ADD-S AUC & ADD(-S) AUC \\ 
		\hline
		&  & 92.3 & 85.4 \\
		$\surd$ &  & 92.7 & 86.6 \\
		$\surd$ & $\surd$ & 94.3 & 89.5 \\
		\hline
	\end{tabular}
\end{table}

\begin{table}
	\centering
	\caption{Ablation studies for differentiable outliers elimination using PoseCNN masks with artificial noises.}
	\label{table:noiseablation}
	\begin{tabular}{cc|cc}
		%\begin{tabular}{m{3.5cm}<{\centering}|m{1cm}<{\centering}|m{1cm}<{\centering}|m{1cm}<{\centering}}
		\hline
		DOE for keypoint & DOE for pose & ADD-S AUC & ADD(-S) AUC \\ 
		\hline
		&  & 92.1 & 84.8 \\
		$\surd$ &  & 92.5 & 86.5 \\
		$\surd$ & $\surd$ & 94.0 & 88.9 \\
		\hline
	\end{tabular}
\end{table}

\section{Implementation Details}
%\subsection{Implementation Details}

\begin{figure}[htbp]
	\centering
	\includegraphics[width=1\linewidth]{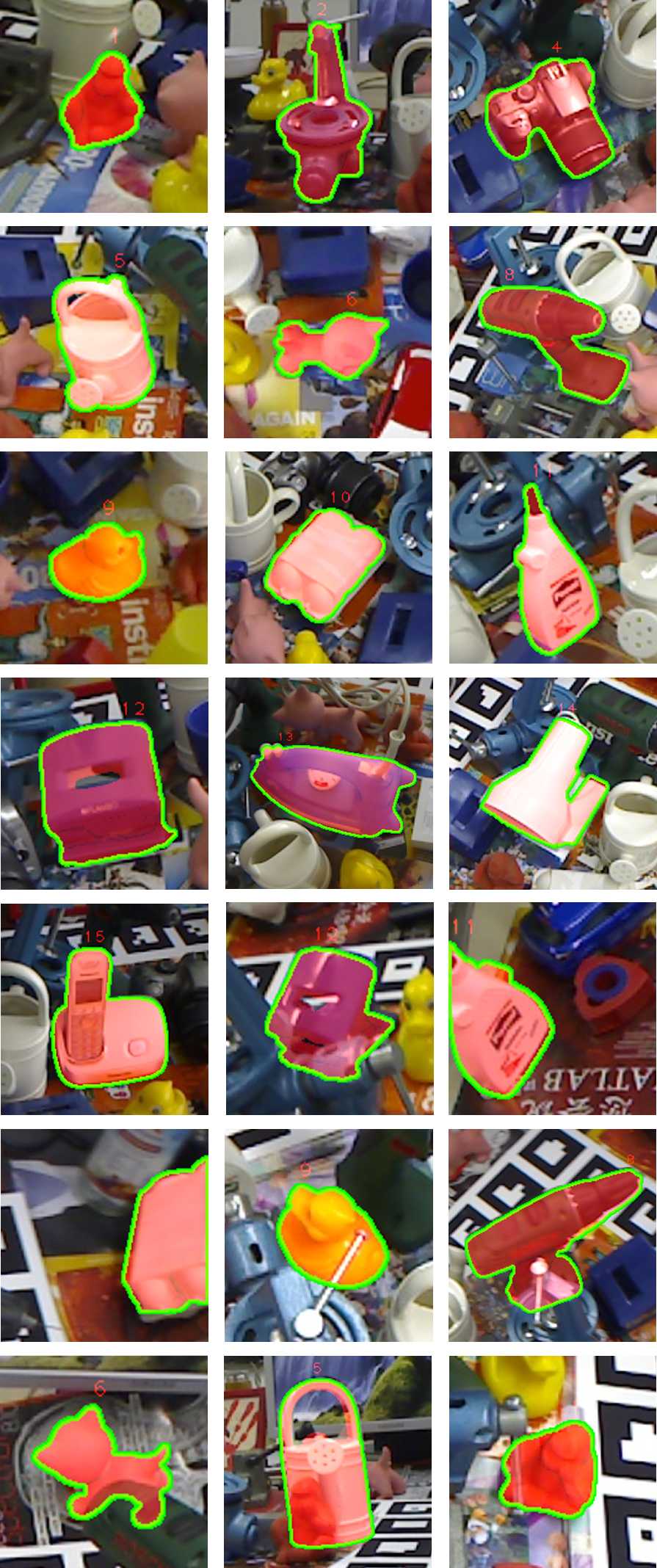}
	\caption{Some visualization results on LineMOD and Occlusion LineMOD dataset. In the pictures, the pose result is shown by projecting the model using the estimated pose. For the convenience of viewing, we also show the contour of the model projection.
	}
	\label{fig:linemod}
	%\label{figurelabel}
	%	\vspace{-0.2cm}
\end{figure}

During implementation, center point and 8 points selected by FPS algorithm are picked up as keypoints following PVNet \cite{peng2019pvnet}. The number of input scene points sampled is 1000 for YCB-Video dataset and 500 for LineMOD dataset. Data augmentation strategies such as random illumination variation are applied to enhance robustness towards brightness and background. 10000 images using the “Cut and Paste” strategy are further synthesized for training on LineMOD dataset as \cite{peng2019pvnet}. The refinement network proposed in DenseFusion \cite{wang2019densefusion} is further employed to iteratively optimize the pose. For YCB-Video dataset, we also employ the ICP algorithm \cite{besl1992method} to improve the performance. The learning rate is set to 1e-4 in pose estimaiton network and 3e-5 in refinement network. The refine margin of ADD(-S) metric is set to 0.013 in YCB-Video dataset and 0.01 in LineMOD dataset. The trade-off parameter $\alpha$ is set to 0.01 and $\beta$ is set to 0.1.

\section{Anylasis for End-to-end Training}

In ablation studies for end-to-end training, we conduct three extra experiments without end-to-end manner for comparision on Occlusion LineMOD dataset. The first experiment extends DPVL \cite{yu20206dof} to 3D and employs SVD 3D-3D estimator. The second and third experiments extend PVNet \cite{peng2019pvnet} to 3D and the second experiment also employs SVD 3D-3D estimator. The third experiment employs RANSAC 3D-3D estimator instead without end-to-end training. Instead of RANSAC, our REDE designs differentiable outliers elimination(DOE) as estimator to obtain deterministic rather than stochastic results, while it is derivable to achieve end-to-end manner. These three results are all worse than REDE (see Table \ref{table:e2e}). Our REDE also outperforms other methods which also employ 3D keypoints but without end-to-end training like PVN3D \cite{he2020pvn3d}, which can prove the importance of end-to-end training.

\section{Anylasis for Differentiable Outliers Elimination}

In ablation studies for differentiable outliers elimination, we conduct three experiments on YCB-Video dataset using different kinds of masks: ground truth mask, PoseCNN mask and PoseCNN mask with artificial noises. For the last mask, we erode PoseCNN mask several times to add artificial noises. Table \ref{table:gtablation}, Table \ref{table:pcablation} and Table \ref{table:noiseablation} report the results respectively. The results for all three types of masks show the effectiveness of our differentiable outliers elimination module. Besides, the improvements using two kinds of inaccurate masks is indeed more obvious, which further proves the robustness of our method to outliers.

\section{More Results}

Some visualization results on LineMOD and Occlusion LineMOD dataset are shown in Fig. \ref{fig:linemod}.

\end{document}